\RequirePackage{fix-cm}
\documentclass[smallextended]{svjour3}       
\smartqed  
\usepackage{graphicx}
\usepackage{amssymb}
\usepackage{amsmath}
\usepackage{url}

\usepackage{breakurl}
\usepackage{xcolor}
\usepackage{graphicx}
\usepackage{subcaption}
\usepackage{multirow}
\usepackage[symbol]{footmisc}
\newcommand{\minisection}[1]{\vspace{0.04in} \noindent {\bf #1}\ \ }

\begin{document}

\title{Main Product Detection with Graph Networks for Fashion}

\renewcommand{\thefootnote}{\fnsymbol{footnote}}
\author{Vacit Oguz Yazici\footnote{the corresponding author}, Longlong Yu, Arnau Ramisa\footnote{work done prior to joining Amazon}, Luis Herranz, Joost van de Weijer}

\authorrunning{Yazici et al.} 

\institute{Vacit Oguz Yazici \at
           Computer Vision Center, Universitat Autonoma de Barcelona, Barcelona, Spain\\
           \email{voyazici@cvc.uab.es}           
       \and
       Longlong Yu \at
       Wide-Eyes Technologies, Barcelona, Spain \\
           \email{longyu@wide-eyes.it}
       \and
         Arnau Ramisa \at
         Amazon Inc., USA \\
         \email{aramisay@amazon.com}
         \and
         Luis Herranz \at
         \email{lherranz@cvc.uab.es}
         \and
         Joost van de Weijer \at
         \email{joost@cvc.uab.es}
}


\maketitle

\begin{abstract}
Computer vision has established a foothold in the online fashion retail industry. Main product detection is a crucial step of vision-based fashion product feed parsing pipelines, focused in identifying the bounding boxes that contain the product being sold in the gallery of images of the product page. The current state-of-the-art approach does not leverage the relations between regions in the image, and treats images of the same product independently, therefore not fully exploiting visual and product contextual information. In this paper we propose a model that incorporates Graph Convolutional Networks (GCN) that jointly represent all detected bounding boxes in the gallery as nodes. We show that the proposed method is better than the state-of-the-art, especially, when we consider the scenario where title-input is missing at inference time and for cross-dataset evaluation, our method outperforms previous approaches by a large margin. 
\end{abstract}

\renewcommand*{\thefootnote}{\arabic{footnote}}
\section{Introduction}
The e-commerce market is growing every year and it is estimated that by 2021 it will make for almost 18\% of the total global retail sales~\cite{ecommerce}. As a consequence, investment in AI technology for fashion that improves the online consumer experience is also increasing~\cite{ecommerce2}. A common problem that AI services companies operating in the fashion industry have, is accurately parsing the feeds with hundreds of thousands of products that the different clients provide as input. Although this task may seem simple at first glance, different patterns of language usage and search engine optimization (SEO) strategies by the merchants (each client can aggregate tens or hundreds of different merchants), combined with visual ambiguity in the images, make achieving industry-grade accuracy very hard. These product feeds often contain fashion products with multiple images depicting a model wearing a complete outfit, and the associated text data like product title, description or category information. 

\begin{figure}
  \includegraphics[width=\linewidth]{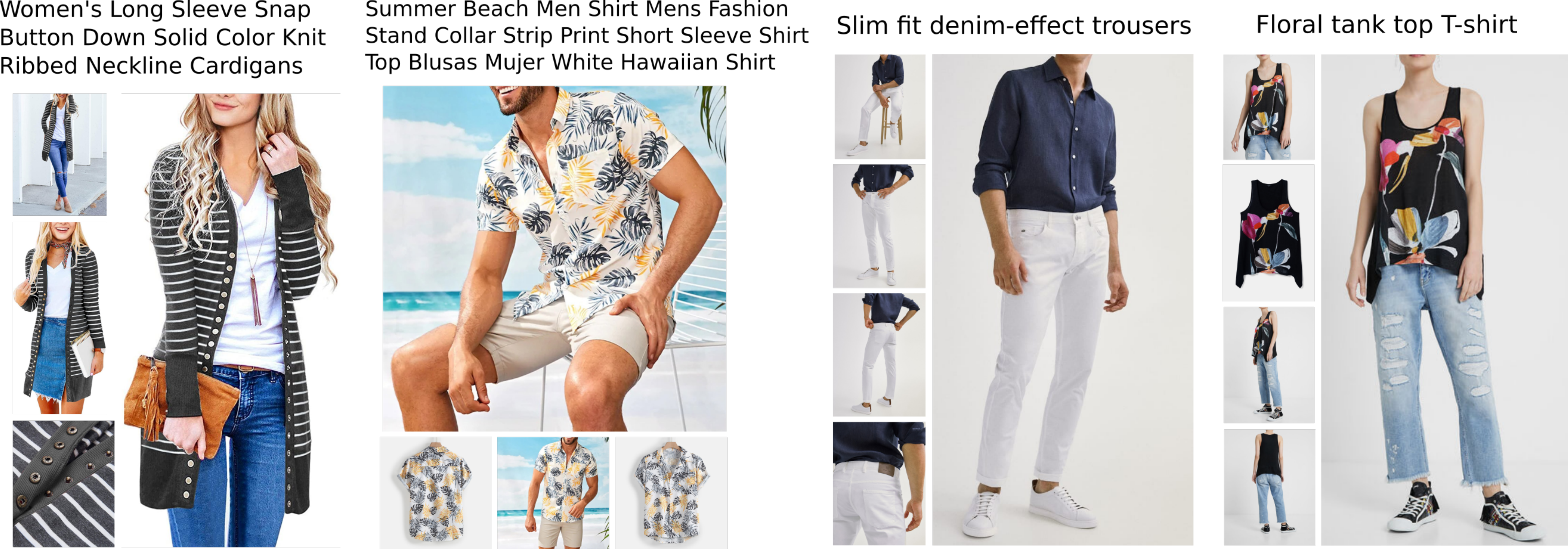}
  \caption{\small Fashion e-commerce sites usually showcase products with a descriptive title and a gallery of images. However, different merchants have different picture and title styles, making it difficult to define generic rules to determine which of the items displayed in the pictures is the one being sold. Therefore, algorithms that can learn this relation are of utmost interest since they would greatly reduce annotation cost.}
  \label{fig:example_merchants}
\end{figure}

More precisely, the task of \emph{main product detection} consists in finding all bounding boxes that contain the product being sold for an input which consists of possibly multiple gallery images combined with a product title (see Figure~\ref{fig:example_merchants}). Finding the main product is a crucial step in many computer vision-based fashion product processing pipelines, as all information derived from the computer vision models that analyze the images will be inaccurate otherwise. Two examples of downstream consequences are wrong category inference and visual search mismatches (e.g. showing a sweater product page when the query image is a skirt). The problem of multi-modal main product detection was defined in~\cite{rubio2017multi}, and is related to visual grounding: a text query (i.e. product title) must be associated with corresponding parts (i.e. bounding box) in a set of gallery images. In their work, they use a contrastive loss in order to learn the representation of positive and negative image-text pairs and treat each bounding box independently, discarding the information of other bounding boxes that belong to the same product. Therefore, the model does not take similarities and dissimilarities between the bounding boxes into account neither during training nor during evaluation. In addition, we introduce the more challenging problem of \emph{gallery-only main product detection}, where at inference time the system has no access to the product title and has to detect the main product only based on the visual information. Although not very common, this setting arises in cases of uninformative product titles, different languages or malformed product feeds, and can lead to costly catastrophic failures if the model cannot recover from it.

\begin{figure}
  \includegraphics[width=\linewidth]{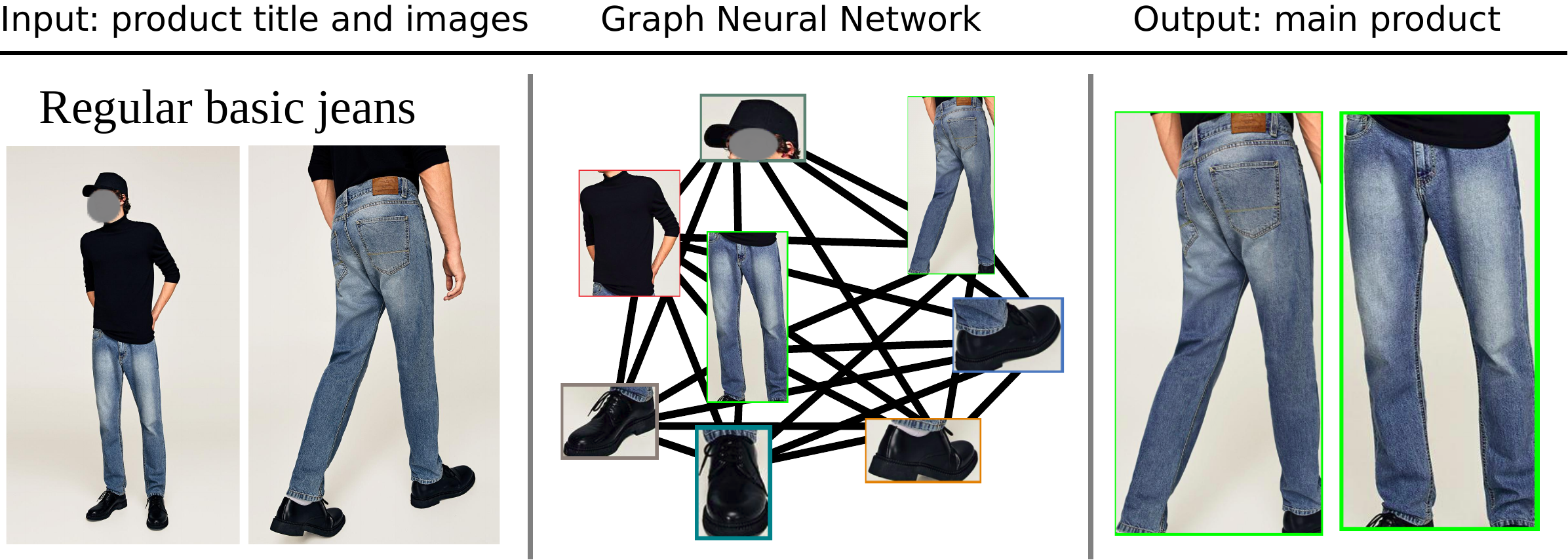}
  \caption{\small Bounding boxes detected in all images of a product are used as nodes in a graph neural network. In this example, inter-image relations are considered for main product detection (jeans).}
  \label{fig:first_figure}
\end{figure}

In our approach, we represent bounding boxes as nodes in a densely connected graph, in which message propagation is realized between all neighbor nodes. In that way, we learn the relation between the images that belong to the same product, exploiting the context provided by all bounding boxes for the prediction (see Figure~\ref{fig:first_figure}). Our model is inspired by the one proposed in~\cite{norcliffe2018learning} for visual question answering. In extensive experiments, we show that taking the context into account leads to improved performance. Especially when considering cross-dataset evaluation where we report a gain of 6-12 points and for the \emph{Gallery-only Main Product Detection} scenario where the text input is missing, where we show that using graphs can result in a gain of up to 50 points when comparing to the same network without graphs. 

This paper is organized as follows, in Section~\ref{sec:related_work}, we introduce the related works that focus on main product detection and incorporate graph convolutional networks for fashion applications. In Section~\ref{sec:method}, we explain our approach and the components of the proposed model in detail. In Section~\ref{sec:experiments}, we describe the experiments that we conduct on the datasets and the results obtained. Finally, in Section~\ref{sec:conclusions}, we summarize our work and draw our main conclusions.

\section{Related Work}
\label{sec:related_work}
The irruption of computer vision and deep learning in the fashion industry has led to many new tasks being proposed to the academic community, such as garment landmark detection~\cite{wang2018attentive,liu2016fashion}, fashion attribute recognition~\cite{ge2019deepfashion2,liu2016deepfashion}, exact product retrieval~\cite{hadi2015buy,kuang2019fashion,ak2018shirt} and compatibility prediction~\cite{cucurull2019context,vasileva2018learning}. In this section we review some of the works most related to ours, namely the ones that use graph convolutional networks or multi-modal embedding learning for fashion-related tasks.

\minisection{Graph Networks for Fashion}
The interest in combining convolutional networks with graph structured data became popular with spectral graph networks proposed in~\cite{bruna2013spectral} and extended by~\cite{kipf2016semi} and~\cite{defferrard2016convolutional}. Velivckovic et al.~\cite{velivckovic2017graph} proposed graph attention networks to exploit masked self-attentional layers to improve the previous methods. Therefore, after the graph networks became popular, new papers emerged which exploit them for traditional computer vision tasks such as image classification
~\cite{chen2019multi,liu2020weakly}, image segmentation~\cite{zhang2019dual}, action recognition~\cite{chen2020graph,yan2018spatial}, anomaly detection~\cite{zhong2019graph} etc.  There are also several works using architectures that include graph neural networks for fashion. Cucurull et al.~\cite{cucurull2019context} propose an apparel compatibility prediction model where clothing items and their pairwise compatibility are represented as a graph, in which vertices are the clothing items and edges connect the items that are compatible. They exploit a graph neural network to predict edge connections in order to find out whether two items are compatible or not. Cui et al.~\cite{cui2019dressing} also propose a model for compatibility prediction with an attention mechanism. In another work~\cite{kuang2019fashion}, the authors use a graph neural network to learn similarities between a query and catalog image in multiple scales, and the similarities are represented by the nodes of a graph that is densely connected. To the best of our knowledge graph neural networks have not been used for main product detection before.

\minisection{Visual-Semantic Joint Embedding for Fashion}
Paired text-image data is very common in the online fashion retail industry, and it has been naturally leveraged to train visual-semantic joint embedding networks. Han et al.~\cite{han2017automatic} propose a concept discovery framework, which automatically identifies attributes derived by jointly modeling image and text. Han et al.~\cite{han2017learning}, employs a bi-LSTM model to jointly learn compatibility relationships among fashion items and a visual-semantic embedding in an end-to-end framework in order to predict compatibility of fashion items and to recommend a fashion item that matches the style of an existing set. Li et al.~\cite{li2017mining}, propose a CNN-RNN model to predict the popularity of a fashion set by fusing text and image features. Liao et al.~\cite{liao2018interpretable}, map fashion features and embeddings of product titles into a joint space in order to obtain meaningful representations and semantic affinities among fashion items. Transformer models have been shown to achieve excellent results in Natural Language Processing, thanks to the abundance of training data. In~\cite{mbastan20tvse}, a large dataset of product title-image pairs is used to train a transformer-based visual semantic embedding, which achieves excellent results at cross-modal retrieval.

\minisection{Main Product Detection}
As mentioned in the previous section, main product detection is a new computer vision task, proposed in~\cite{rubio2017multi}. Their proposed model has 3 main components which are the contrastive loss, the classification losses, and the word2vec model~\cite{mikolov2013distributed} that extracts the product title embeddings. The contrastive loss is used for positive and negative image-text pairs. Auxiliary classification losses for image and text are used to improve training stability and performance. To train a word2vec model, they concatenate all the available text fields in their feeds, then compute 100-dimensional descriptors for each word appearing more than 5 times. Finally, they average the descriptors to get the product title embeddings. They treat each image independently during training and evaluation which means that they do not take the relation between images that belong to a same product into account. For the rest of the paper, we will denominate this paper as \emph{Contrastive model}.

\begin{figure*}[!tb]
\begin{center}
\includegraphics[width=\textwidth]{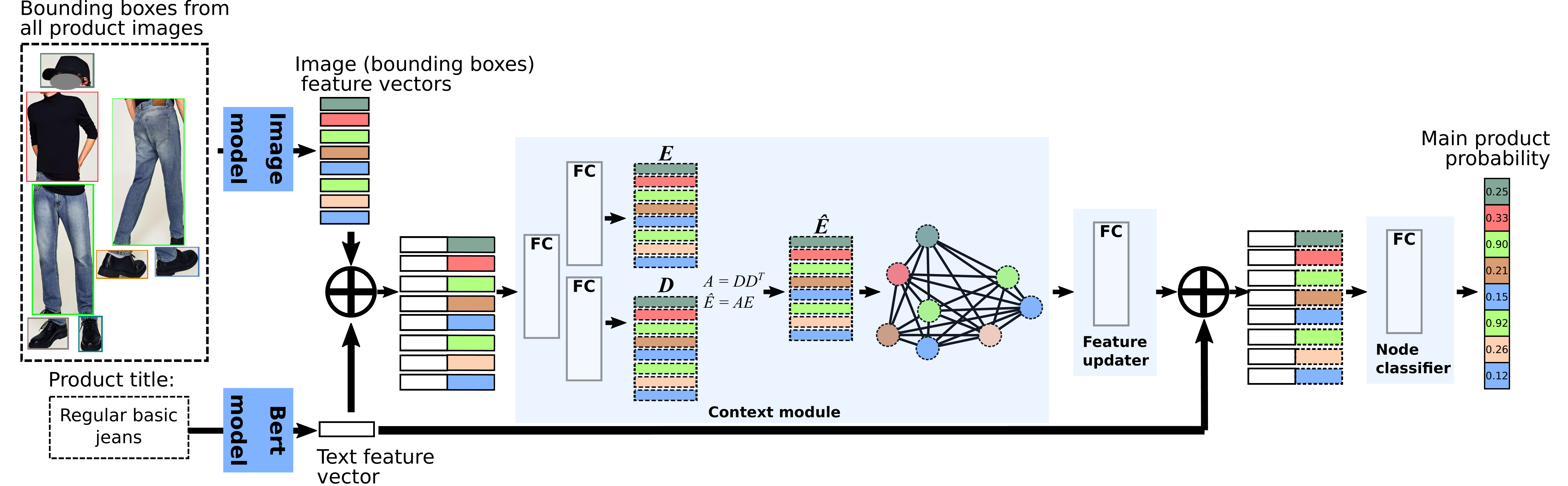}
\caption{The architecture of the model. The image features for bounding boxes from all product images are concatenated with the product title embedding. These are then used as nodes of the graph. The probability that they are the main product is estimated for each one. We also display the other variants of our model in Figure~\ref{fig:other_variants}.} 
\label{fig:framework}
\end{center}
\end{figure*}

\section{Method}
\label{sec:method}
Main product detection deals with associating correct parts of images (bounding boxes) with the given product title. As discussed before, prior work~\cite{rubio2017multi} considers the bounding boxes separately to decide on which of them correspondent to the product title. However, it is likely that a good view of the product in one gallery image should be able to help the algorithm identify the main product in other images where it is featured less prominently. Therefore, we take a more holistic view to the problem and we want the algorithm to consider all parts in all the gallery images simultaneously.

Figure~\ref{fig:framework} shows the architecture of our proposed model, which consists of five parts: image model, BERT (text) model~\cite{devlin2018bert}, context module, feature updater and node classifier. The input for the BERT model are product titles, while the input for the image model are image crops corresponding to the bounding boxes. The graph in the context module is densely connected, and the nodes represent the bounding boxes found in the product gallery images. Let $G = \{{V}, \mathcal{E}, {A} \}$ be an undirected graph with self-loops, where $\mathcal{E}$ and ${V} \in \mathbb{R}^{N \times d}$ represent the edges and nodes respectively, and ${A} \in \mathbb{R}^{N \times N}$ the corresponding adjacency matrix. $N$ and $d$ are number of nodes and dimension of node features respectively. The idea is to learn the relations between the nodes (bounding boxes) given the title and help classify them correctly.

\minisection{Image model} The Image model is a ResNet-34~\cite{he2016deep} convolutional neural network that extracts features for each given bounding box. Activations from \textit{layer4} are average pooled (512 dimensions) and fed to the next stage of the architecture. The model is initialized with pre-trained ImageNet weights.

\minisection{BERT model} In order to extract sentence embeddings for each title, we use a pre-trained BERT model~\cite{devlin2018bert}. For the dataset with the product titles in English, we use the \textit{bert-base-uncased} BERT model, and for the one with the product titles in Turkish we use the \textit{bert-multilingual-cased}\footnote{https://github.com/google-research/bert} model. We apply the BERT tokenizer which splits strings in sub-word token strings that convert them to indexes according to mappings in its vocabulary. The model outputs an embedding for each token. To extract the sentence embeddings, we use the average max pooling method (i.e. concatenation of average pooled and max pooled tokens into one vector). Since the dimensionality of the BERT models is 768, after concatenation it doubles in size and becomes 1536, so we add an extra fully connected layer to reduce the dimensionality to 512.

\begin{figure}[!tb]
  \centering
\begin{subfigure}{.49\textwidth}
  \centering
  \includegraphics[width=.7\linewidth]{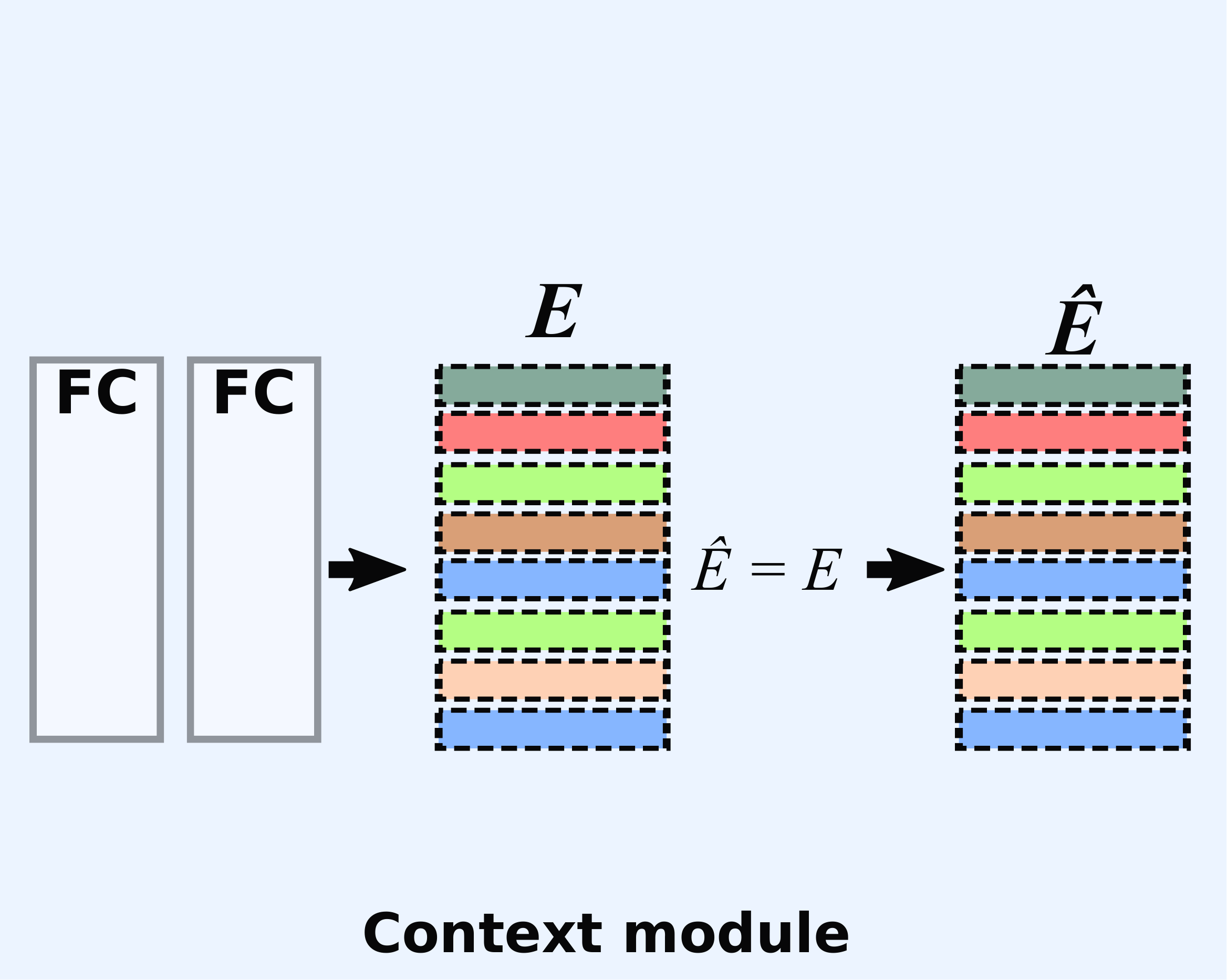}
  \caption{NG}
  \label{fig:sfig1}
\end{subfigure}%
\begin{subfigure}{.49\textwidth}
  \centering
  \includegraphics[width=.9\linewidth]{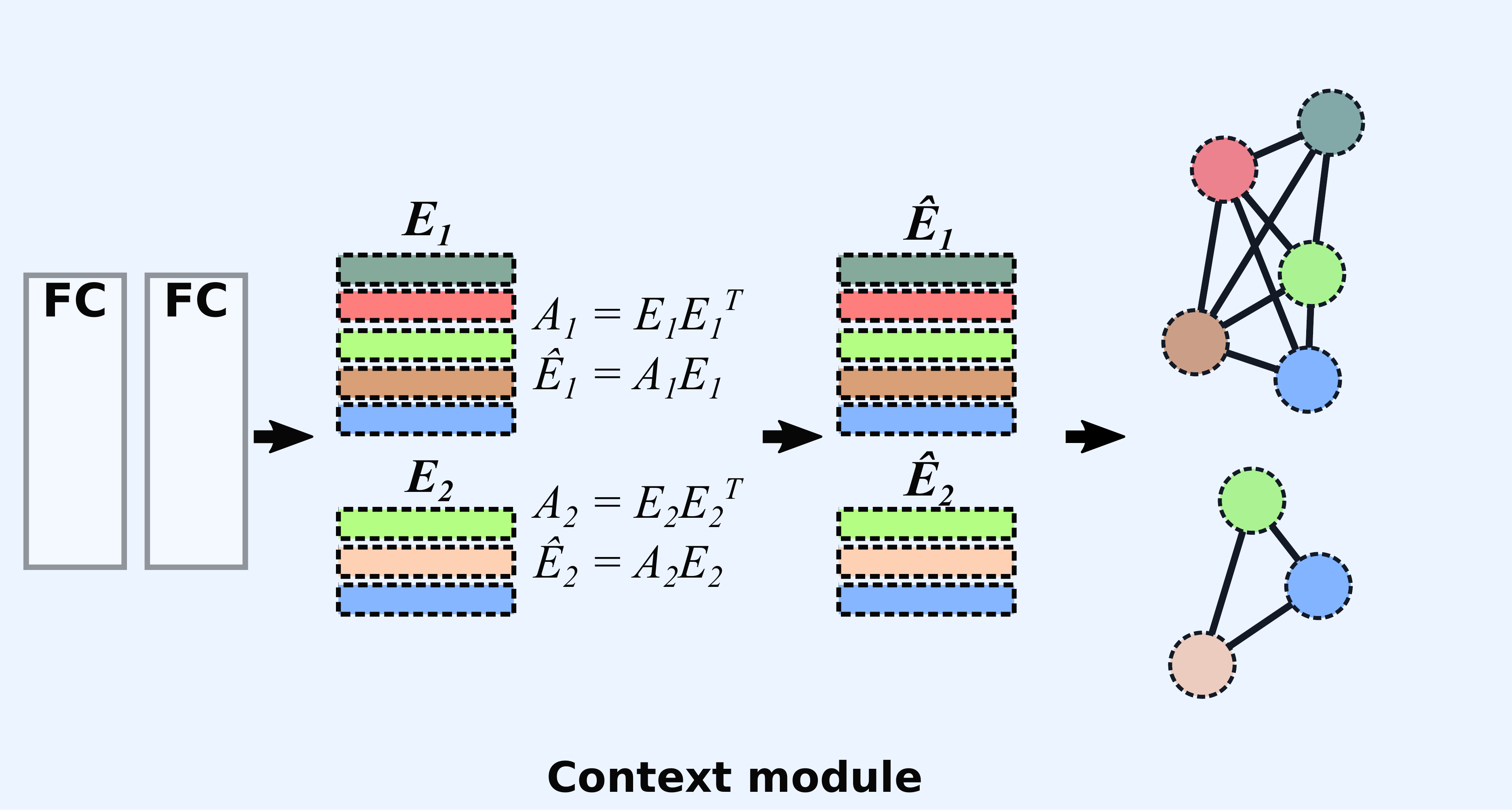}
  \caption{ICFS}
  \label{fig:sfig2}
\end{subfigure}
\begin{subfigure}{.49\textwidth}
  \centering
  \includegraphics[width=.9\linewidth]{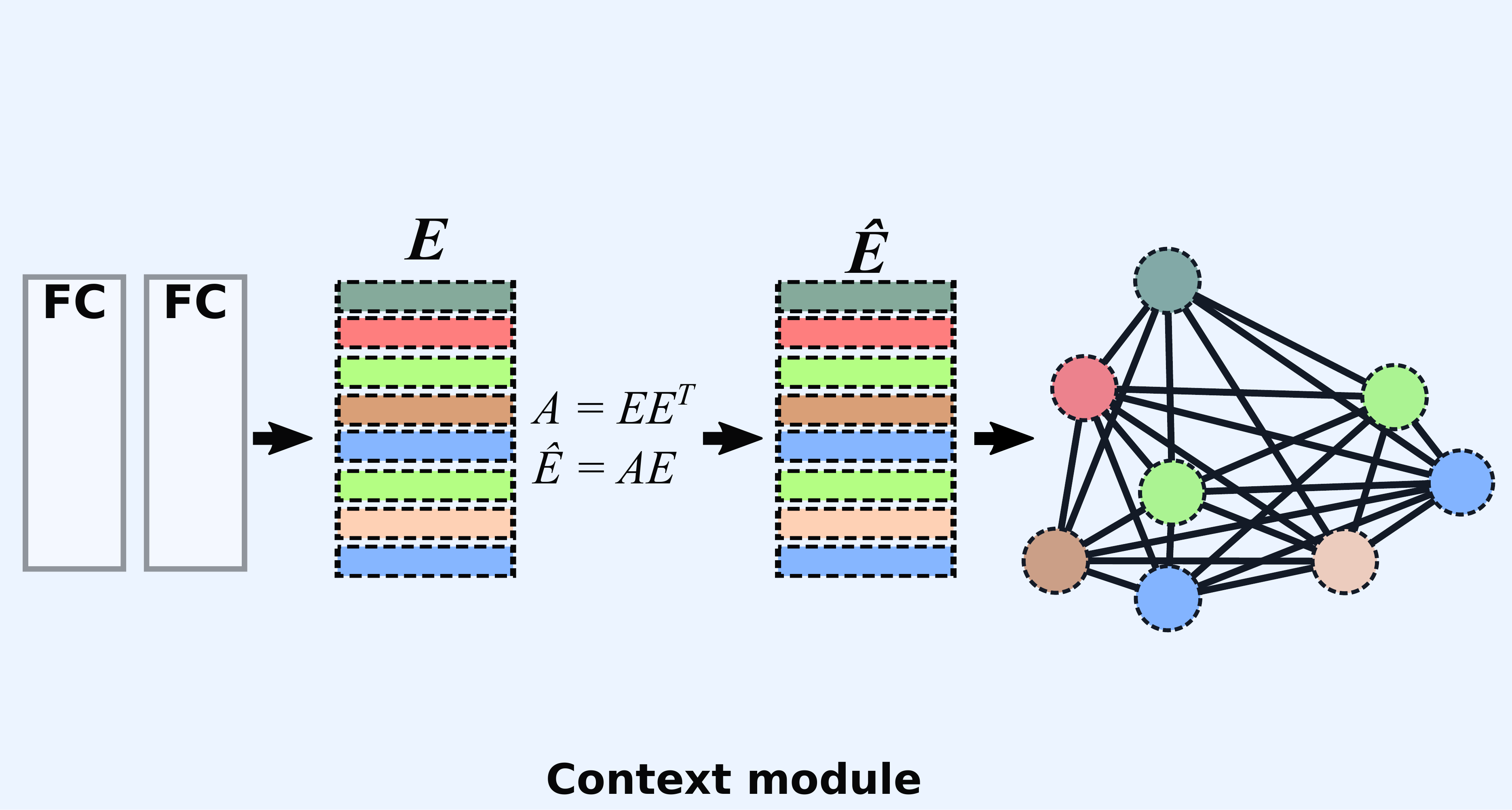}
  \caption{PCFS}
  \label{fig:sfig3}
\end{subfigure}
\begin{subfigure}{.49\textwidth}
  \centering
  \includegraphics[width=.9\linewidth]{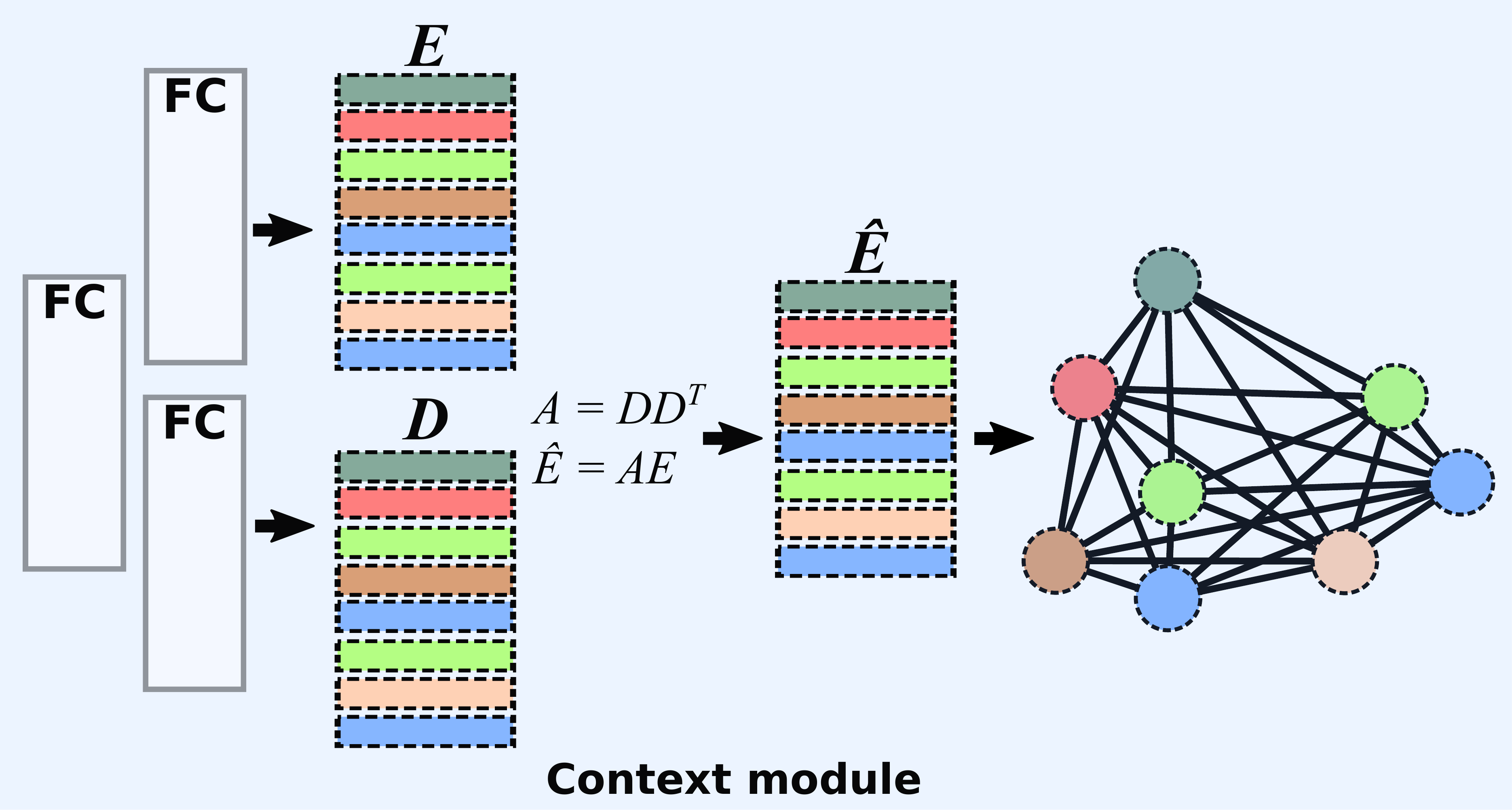}
  \caption{PDFS}
  \label{fig:sfig4}
\end{subfigure}
\caption{The context modules of baseline and variants of our model. (a) In the no-graph model (NG) there is no graph to represent the bounding boxes as nodes as there is no interaction between the boxes. (b) In the ICFS (Instance Coupled Feature Similarity) we represent each product image as a graph. (c) In the PCFS (Product Coupled Feature Similarity) graph model, the same features are used to get the adjacency matrix and updated features. (d) The PDFS (Product Decoupled Feature Similarity) graph model, decouples the update of node feature and calculation of adjacency matrix.}
\label{fig:other_variants}
\end{figure}

\minisection{Context module} The main novelty of our work is the introduction of the graph network within the context module. The graph network models the interaction between the various items shown in the image gallery and the product title (see Figure~\ref{fig:other_variants}). Since the proposed graph topology is densely connected, the message passing between the nodes cannot be a simple sum of neighbor node features, as it will make all node features equal in the next layer. Therefore, we use the graph learner architecture proposed in~\cite{norcliffe2018learning}, that learns the adjacency matrix for the message passing. As mentioned before, one node corresponds to each bounding box, and the edges connect every pair of nodes. We build the node features by concatenating bounding box and title embeddings, represented as $[{f}_n, {t}]$ for bounding box feature $f_n$ and the title embedding $t$, and input them to the graph learner $F$, which consists of two fully connected layers with ReLU activation:
\begin{equation}
{e}_n = F([ {f}_n, {t}])
\end{equation}
The dimensionality of $[{f}_n, {t}]$ is 1024 (512 + 512), but it is reduced back to 512 after the first layer. All $N$ output features ${e}_n$ are stacked into a matrix $E \in \mathbb{R}^{N \times P}$, where $P$ is the dimension of the concatenated features, we compute the adjacency matrix with the following equation:
\begin{equation}
A = E{E}^T
\end{equation}
which is defined as a fully connected adjacency matrix. This is not a problem computationally since the number of nodes per product is low in our problem (we will show the statistics in the datasets section). The adjacency matrix is then used for message passing before the node feature update:
\begin{equation}
\hat{E} = AE
\end{equation}
We denominate this model as Coupled Feature Similarity (CFS). In CFS, $E$ is used for obtaining the adjacency matrix and also as input features ($\hat{E}$) for the graph. Therefore, calculation of the adjacency and node feature update are coupled. However, we observed that using the same features $E$ for these two purposes (i.e. pairwise similarity and node representation) may be limiting, so we propose to increase the flexibility of the model by allowing it to decouple them and learn specific representations for each of those purposes. Therefore, we test a variant of our model in which, instead of obtaining the adjacency matrix as a product of $E$ and $E^{T}$, an additional fully connected layer (head) after the context module is used to obtain matrix $D \in \mathbb{R}^{N \times D}$ (see Figure~\ref{fig:framework}), which is subsequently used for message passing:

\begin{equation}
\begin{aligned}
& {e}_n, {d}_n = F([ {f}_n, {t}]) \\
& A = D{D}^T \\
& \hat{E} = AE
\end{aligned}
\end{equation}
As before, all output features ${d}_n$ are stacked into a matrix $D$. This formulation allows us to directly learn the adjacency matrix instead of extracting it from the node features. Since this model decouples the update of node feature and calculation of adjacency matrix, we denominate it as Decoupled Feature Similarity (DFS).

\begin{figure}[!tb]
 \begin{center}
  \includegraphics[width=0.79\linewidth]{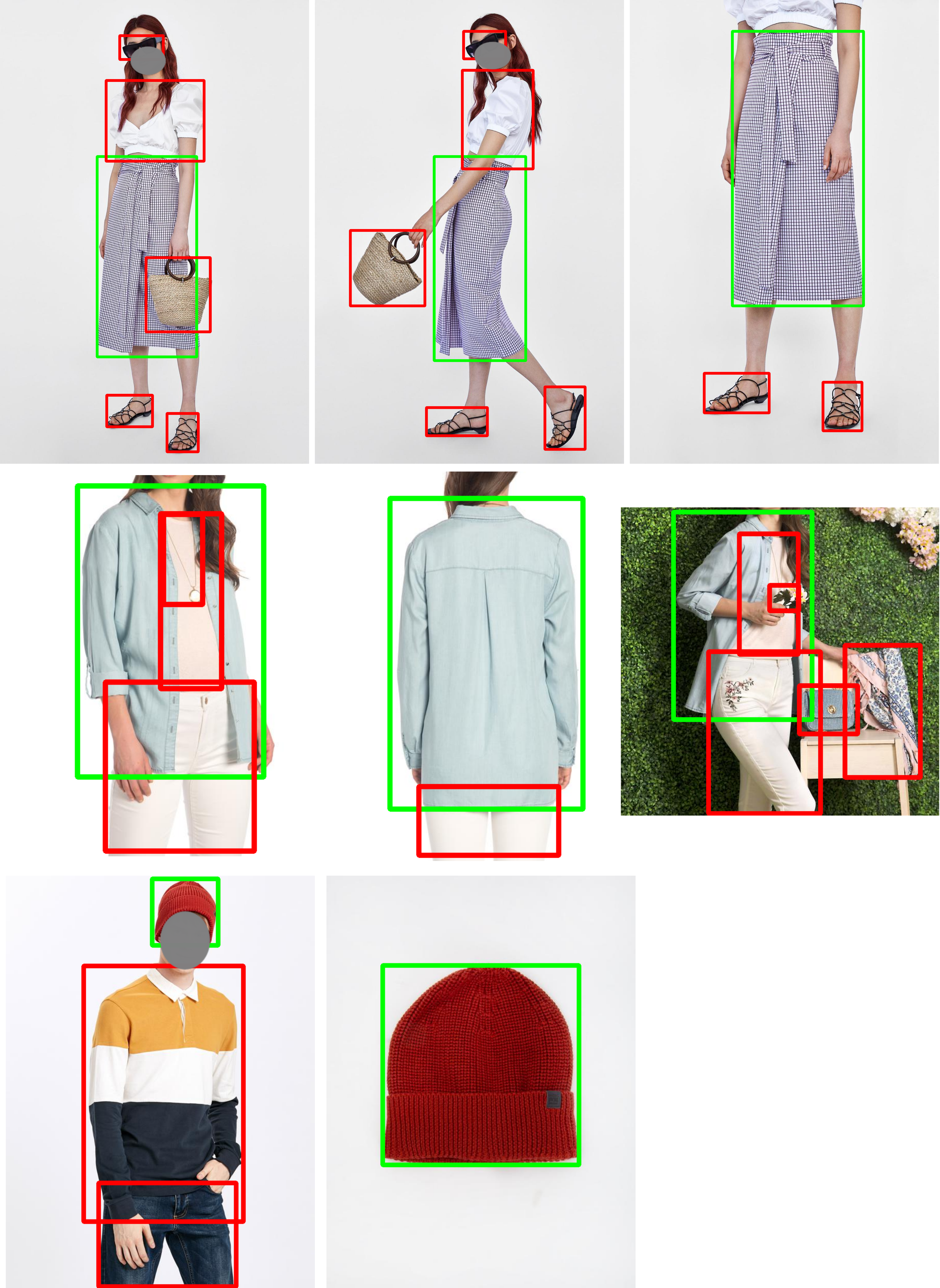}
  \caption{\small Some multi-language example products from the dataset. The main bounding boxes are drawn in green. The titles of the products are: \textit{Checked wrap skirt}, \textit{Kadın gömlek}(\textit{Woman shirt}) and \textit{Triko bere}(\textit{Knit beanie}) respectively. All the bounding boxes are computed with a fashion product detector pre-training.}
   \label{fig:dataset}
 \end{center}
\end{figure}

The baseline and variants of our model are displayed in Figure~\ref{fig:other_variants}. As can be seen, we consider two setups for the CFS models: Instance Coupled Feature Similarity (ICFS) and Product Coupled Feature Similarity (PCFS). In the ICFS, we represent each product image as a graph. Because of this, and in contrast with the baseline NG model, it is allowed to take into account the context provided by the negative bounding boxes in the same image during training and evaluation. However, it does not fully exploit the relation between all bounding boxes since they are not densely connected as in the PCFS model. We do consider the connections between all bounding boxes in all images in the PCFS model. 

\minisection{Feature updater} The feature updater part consists of one fully connected layer and a leaky ReLU activation. We have also added these layers to the no-graph  baseline model (NG) to allow for a fair comparison with the graph-based models (to ensure that they have a comparable capacity as our proposed methods).

\minisection{Node classifier} The input of the node classifier is the concatenation of the original BERT embeddings and the output node features. It consists of a single fully connected layer to reduce the dimensionality to 2 (node active or inactive), and it is followed by the binary cross entropy loss during training.

\section{Experiments}
\label{sec:experiments}
\subsection{Datasets}
We evaluated the proposed methods on two datasets whose statistics can be seen in Tables~\ref{tab:dataset_stats} \& ~\ref{tab:dataset_stats2}. We crawled each of the datasets from a different e-commerce website. We collected information related to title, description, attribute information and product images, on which we ran a fashion product detector to get bounding boxes. Finally, we used human annotators to label the ground truth main bounding boxes for each product gallery. We split the datasets and allocate 75\%, 5\%, 20\% for training, validation and test sets respectively. Some example products can be seen in Figure~\ref{fig:dataset}\footnote{Contact the corresponding author to obtain the original urls of images, product titles, splits and main bounding box information.}.

As an extra experiment, we evaluate our models on the main bounding box detection dataset (MBBDD) which was made public by~\cite{rubio2017multi}. Due to the significant amount of time has passed by since the dataset was first made public, we were able to recover only a subset of the dataset. Out of total 458,700, we retrieved 91,550 products. The number of images per product is 1 and the number of bounding boxes per image is 2.37. We use 77,820 products for the training and validation and the rest of them as a test set. Instead of using bounding box proposals, we use the same fashion product detector that we used for our datasets to get bounding boxes. The rest of the details about the dataset can be found in~\cite{rubio2017multi}.

\begin{table}[!tb]
\centering
\caption{Dataset statistics. BBs denotes bounding boxes.}
\scalebox{0.6}{
\begin{tabular}{cc|c|c|c|c|c|c|c|c|c|c|c}
\multirow{2}{*}{Datasets} & \multirow{2}{*}{Lang.} & \multicolumn{9}{c|}{Categories} & \multirow{2}{*}{Images/product}& \multirow{2}{*}{BBs/image} \\
& & \multicolumn{1}{c}{\makebox[1em][l]{\rotatebox{45}{accessory}}} & \multicolumn{1}{c}{\makebox[1em][l]{\rotatebox{45}{bags}}} & \multicolumn{1}{c}{\makebox[1em][l]{\rotatebox{45}{bottom}}} & \multicolumn{1}{c}{\makebox[1em][l]{\rotatebox{45}{swim}}} & \multicolumn{1}{c}{\makebox[1em][l]{\rotatebox{45}{one-piece}}} & \multicolumn{1}{c}{\makebox[1em][l]{\rotatebox{45}{outerwear}}} & \multicolumn{1}{c}{\makebox[1em][l]{\rotatebox{45}{shoes}}} & \multicolumn{1}{c}{\makebox[1em][l]{\rotatebox{45}{sweaters}}} & \multicolumn{1}{c|}{\makebox[1em][l]{\rotatebox{45}{top}}}  & & \\\hline
1                      & English                   & 236       & 440  & 4711   & -    & 1820      & 2972      & 441   & 2474     & 6424 & 4.40                                                               & 2.40                                                                     \\
2                      & Turkish                   & 2220      & 556  & 5183   & 811  & 1263      & 1190      & 1244  & 3521     & 6290 & 2.46                                                               & 2.73 
\end{tabular}
}
\label{tab:dataset_stats}
\end{table}

\begin{table}[!tb]
\centering
\caption{Number of images with $M$ bounding boxes.}
\scalebox{0.6}{
\begin{tabular}{c|c|ccccccccccc}
Datasets           &                                                                                     & $M\!=\!1$     & $M\!=\!2$     & $M\!=\!3$     & $M\!=\!4$     & $M\!=\!5$    & $M\!=\!6$    & $M\!\!=\!\!7$   & $M\!=\!8$    & $M\!=\!9$    & $M\!=\!10$   & $M\!>\!10$ \\ \hline
\multirow{2}{*}{1} & Per image   & 33747 & 17590 & 10945 & 16111 & 5750 & 1410 & 283 & 32   & 8    & 3    & 1                \\
                   & Per product& 1118  & 992   & 1085  & 1578  & 977  & 790  & 738 & 878  & 974  & 1038 & 9350             \\ \hline
\multirow{2}{*}{2} & Per image & 17334 & 12170 & 6955  & 10360 & 4841 & 1985 & 660 & 264  & 92   & 44   & 92               \\
                   & Per product & 1980  & 4849  & 1507  & 1439  & 591  & 1386 & 865 & 2052 & 1057 & 1263 & 5289     
\end{tabular}}
\label{tab:dataset_stats2}
\end{table}

\subsection{Evaluation metrics}
We consider the product accuracy for a single product to be 1 if all positive (product being sold) and negative (other parts of the outfit) bounding boxes are classified correctly, and 0 otherwise. Then all scores for all test images are averaged to get the final score. We deem the product accuracy metric to be the most important indicator for a main product detection system. As we explained before, one wrong bounding box classification might cause visual search mismatches in queries related to the product. Therefore, it is crucial to classify all bounding boxes of a product correctly to avoid such problems. We also consider the precision@1, recall@1 and mAP metrics. For the graph based models, we use the classification scores to rank the nodes of a product. For the contrastive model, we use the distances between image features and title embeddings.

\subsection{Network training}
We implemented our architecture using the PyTorch framework~\cite{NEURIPS2019_9015} and Deep Graph Library~\cite{wang2019dgl}. The Adam optimizer is chosen for the training. We use learning rate $10^{-4}$ and $3\times10^{-6}$ for the image and BERT models respectively. For the remaining parts of the model, the learning rate is $10^{-4}$. The batch size is 6 and each batch sample is a graph of nodes that represent the bounding boxes that belong to same products. In all experiments, we train the models for 25 epochs, and the snapshot that yields the best accuracy on the validation set is evaluated on the test set for the reported results. 
For the contrastive model we use batch size of 32 and train for 35 epochs. This was done to obtain competitive results compared to our methods. In the evaluation, we choose a node as a positive node if the probability of the final score is higher than $0.5$. For [26], we set the margin hyper-parameter of the contrastive loss to $0.5$ for training. During evaluation, we accept as main product the detections that have a cosine distance lower than $0.1$ with the product title embedding. Both values were selected by cross-validation.

\begin{figure*}[!tb]
\begin{center}
\includegraphics[width=0.8\textwidth]{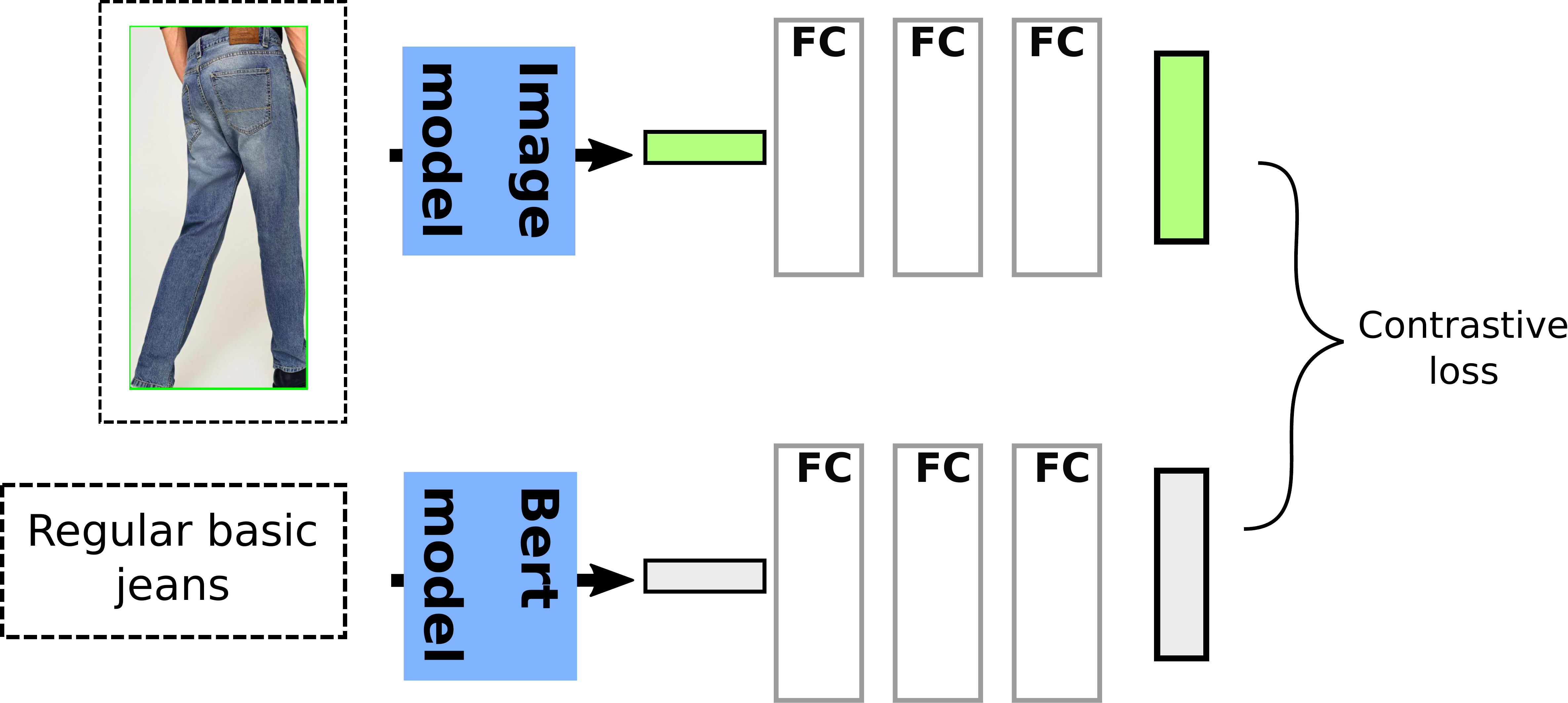}
\caption{The architecture of the contrastive model.} 
\label{fig:framework_contrastive}
\end{center}
\end{figure*}

\subsection{Comparison with the Baseline models}
In the initial experiments, we compare the proposed approach with a no-graph (NG) model, which contains the same layers as the proposed model (see Figures~\ref{fig:framework} and~\ref{fig:other_variants}). The only difference is that the adjacency matrix is not used, as there is no node feature update step in the NG model. Therefore, bounding boxes cannot interact, and each decision is computed independently from the others.

Our second baseline model is the \emph{Contrastive model}~\cite{rubio2017multi}, where the authors propose to map the image and text embeddings into a common space, and reduce the distances between positive bounding boxes and their titles with a contrastive loss, as well as including additional auxiliary losses for bounding box and text classification. To make the models comparable, we make sure that the image and text branches have the same architectures, we include the extra fully connected layers in the other parts of the model, and  remove every loss apart from the contrastive loss. Since we cannot concatenate features and embeddings as we do in our proposed model, we create two branches for image features and text embeddings after the image and BERT models (see Figure~\ref{fig:framework_contrastive}). Then, we compare our graph-based approaches: ICFS, PCFS nad PDFS. To make the comparison fair with the other graph-based models, we evaluate the ICFS model by checking the image score (which is 1 if all bounding boxes of an image are classified correctly 0 otherwise) and assigning 1 to product score if all image scores are 1. 

\begin{table}[!tb]
\centering
\caption{Performance comparison of the baselines and graph-based approaches}
\begin{tabular}{ccc|cccc}
Train               & Test               & Models      & P@1           & R@1           & mAP           & Prod. acc.    \\ \hline
\multirow{10}{*}{1} & \multirow{5}{*}{1} & Contrastive & 98.7          & 32.2          & 99.1          & 81.0          \\
                    &                    & NG          & 99.3          & 32.4          & 99.5          & 87.1          \\
                    &                    & ICFS        & \textbf{99.3} & \textbf{32.4} & \textbf{99.5} & 88.1          \\
                    &                    & PCFS        & 98.6          & 31.6          & 99.1          & 87.6          \\
                    &                    & PDFS        & 99.1          & 32.1          & 99.4          & \textbf{89.1} \\ \cline{2-7} 
                    & \multirow{5}{*}{2} & Contrastive & 97.1          & 44.8          & 98.1          & 84.3          \\
                    &                    & NG          & \textbf{98.1} & \textbf{45.2} & 98.4          & 84.7          \\
                    &                    & ICFS        & 97.7          & 45.0          & 98.1          & 87.8          \\
                    &                    & PCFS        & 96.5          & 44.7          & 97.6          & 87.1          \\
                    &                    & PDFS        & 97.7          & 45.1          & \textbf{98.5} & \textbf{90.3} \\ \hline
\multirow{10}{*}{2} & \multirow{5}{*}{1} & Contrastive & 92.7          & 30.5          & 93.4          & 41.1          \\
                    &                    & NG          & 95.4          & 31.0          & 93.9          & 43.2          \\
                    &                    & ICFS        & 96.0          & 31.1          & 94.7          & 53.7          \\
                    &                    & PCFS        & \textbf{96.1} & \textbf{31.4} & \textbf{95.8} & 51.0          \\
                    &                    & PDFS        & 93.1          & 30.4          & 94.5          & \textbf{55.7} \\ \cline{2-7} 
                    & \multirow{5}{*}{2} & Contrastive & 99.2          & 45.8          & 99.4          & 92.0          \\
                    &                    & NG          & 99.6          & 46.0          & 99.6          & 94.7          \\
                    &                    & ICFS        & \textbf{99.7} & \textbf{46.1} & \textbf{99.6} & 94.5          \\
                    &                    & PCFS        & 99.5          & 46.0          & 99.6          & 94.9          \\
                    &                    & PDFS        & 99.5          & 46.0          & \textbf{99.6} & \textbf{95.6}
\end{tabular}
\label{tab:results1}
\end{table}

\begin{figure}[!tb]
  \centering
\begin{subfigure}{.5\textwidth}
  \centering
  \includegraphics[width=.95\linewidth]{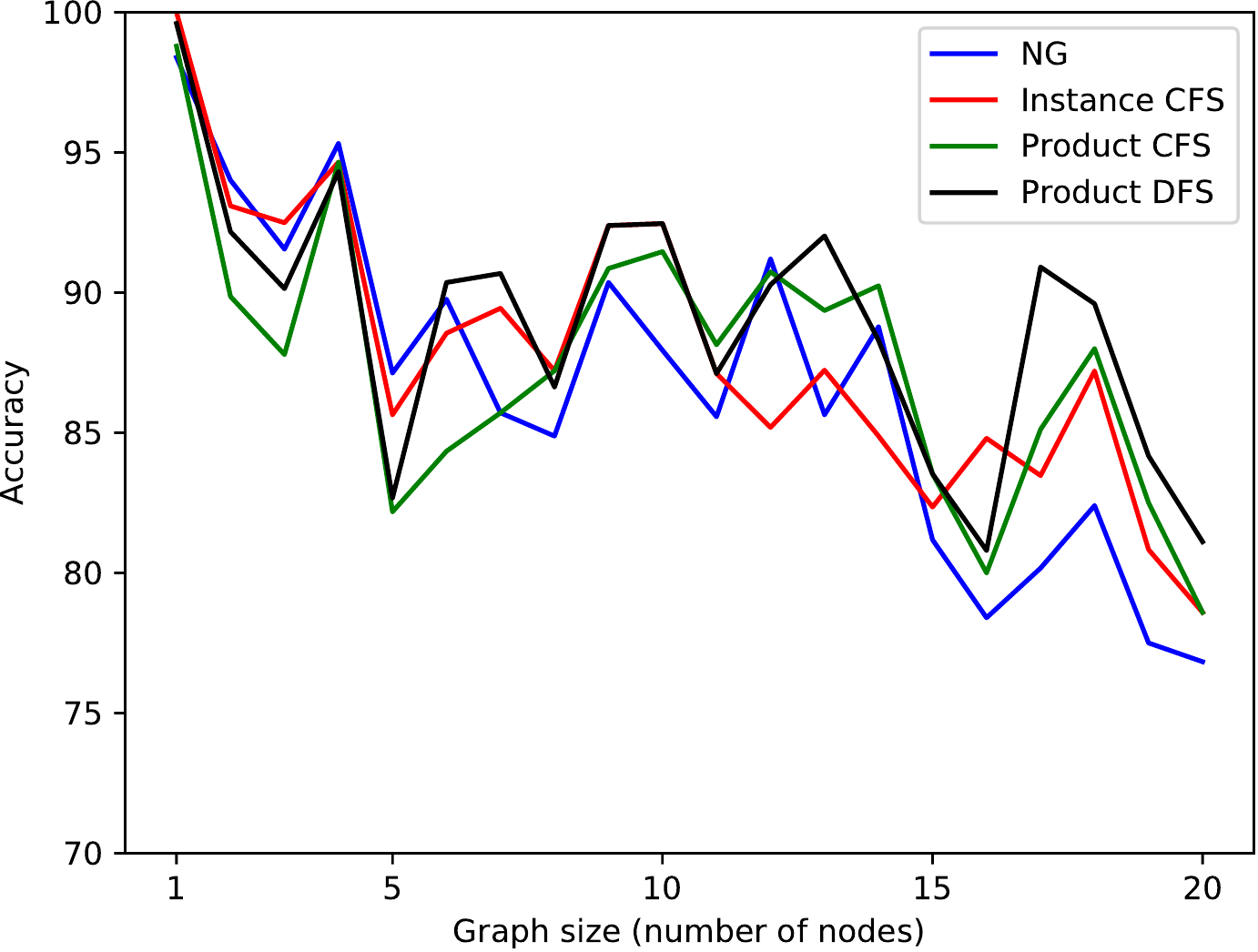}
  \caption{Main Product Detection}
  \label{fig:curve1}
\end{subfigure}%
\begin{subfigure}{.5\textwidth}
  \centering
  \includegraphics[width=.95\linewidth]{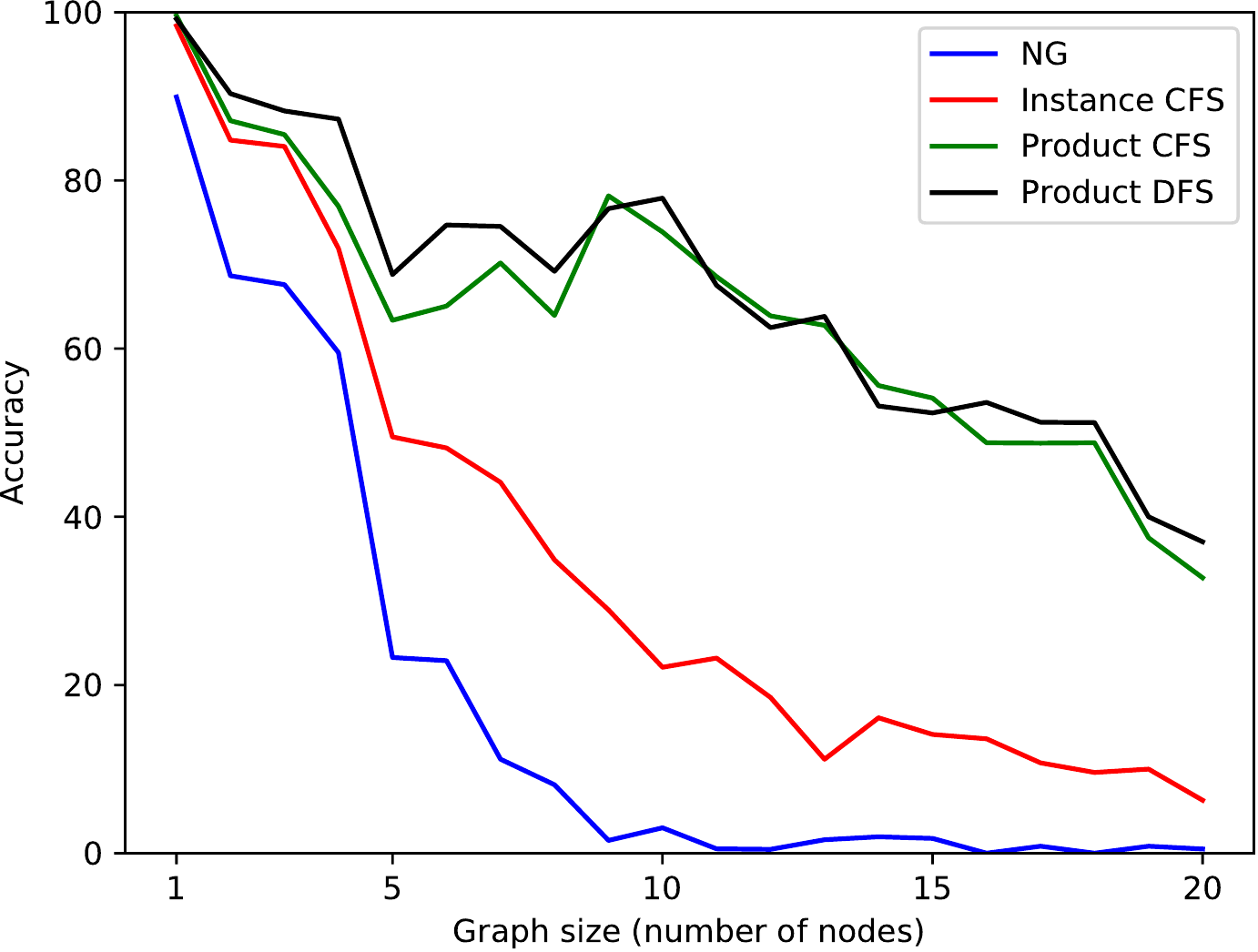}
  \caption{Gallery-only Main Product Detection}
  \label{fig:curve2}
\end{subfigure}
\caption{\small Comparison of accuracies of different models with changing graph sizes on the dataset 1. Our proposed method especially improves results when the gallery of images contains many bounding boxes.}
\label{fig:curve}
\end{figure}

The results are summarized in Table~\ref{tab:results1}. We first focus on the in-dataset evaluation, referring to the results where train and test set originate form the same dataset. As can be seen the graph-based methods outperform the baselines in the product accuracy metric by a significant margin. Especially our PDFS model manages to obtain good results in the product accuracy metric, outperforming the other graph-based methods and the NG baseline. Since the average graph size is bigger for dataset 1 (see Tables~\ref{tab:dataset_stats} and~\ref{tab:dataset_stats2}), the gain with the graph-based models is higher for the dataset 1. Precision@1 and recall@1 metrics yielded by the graph-based and baseline models are comparable, because it is relatively easy task to sort the bounding boxes by similarity since the number of nodes per product is low. However, in most of the metrics, our graph-based models obtain better scores. The change in performance with the graph size is further analyzed in Figure~\ref{fig:curve1}. All the graphs whose size is bigger than 20, are represented as their size is $20$ in the figure. As expected, graph-based approaches can handle larger graphs better than the non-graph based approaches since it gets harder to classify all the nodes correctly when the number of nodes increases in the absence of context.  

We also do cross-dataset evaluation to assess the generalization ability of the models. For the cross-dataset evaluation, we translate the titles from English to Turkish and from Turkish to English by using a Google Translator API. In this case the gains because of the graph-model are more pronounced, especially when evaluating the model trained on dataset 2 on dataset 1, where results increase from 43.2\% (NG) to 55.7\% (PDFS), showing that the graph-based methods generalize better to new data.

\begin{figure}[t]
  \centering
 \begin{subfigure}{.9\textwidth}
  \centering
  \includegraphics[width=1\textwidth]{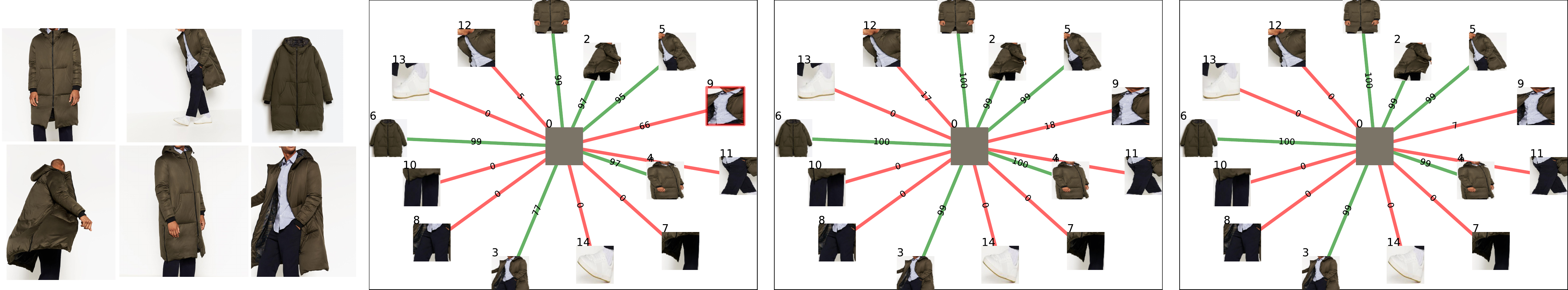}
    \caption{\textit{Feather coat}}
  \label{fig:compare1}
\end{subfigure}%
\\
\begin{subfigure}{.9\textwidth}
  \centering
  \includegraphics[width=1\textwidth]{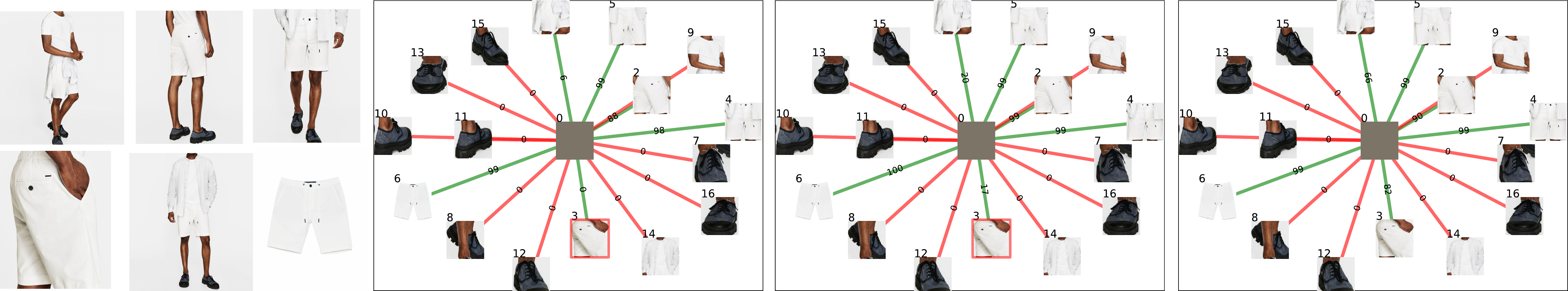}
    \caption{\textit{Chino bermuda shorts}}
  \label{fig:compare2}
\end{subfigure}%
\\
\caption{ Qualitative evaluation of NG, PCFS and PDFS models respectively. The product titles are written under the subfigure. The gray nodes in the middle are the title nodes which are presented for demonstration purposed. The scores on the edges are the classification scores of the connected nodes. The green and red edges represent positive and negative nodes respectively. The nodes that are bounded by a red box are the wrong classifications. The superiority of the graph based models are more apparent in case of larger graph sizes (see also Figure~\ref{fig:curve2}) (best viewed in color).}
\label{fig:compare}
\end{figure}

\begin{figure}[t]
 \begin{center}
  \includegraphics[width=\linewidth]{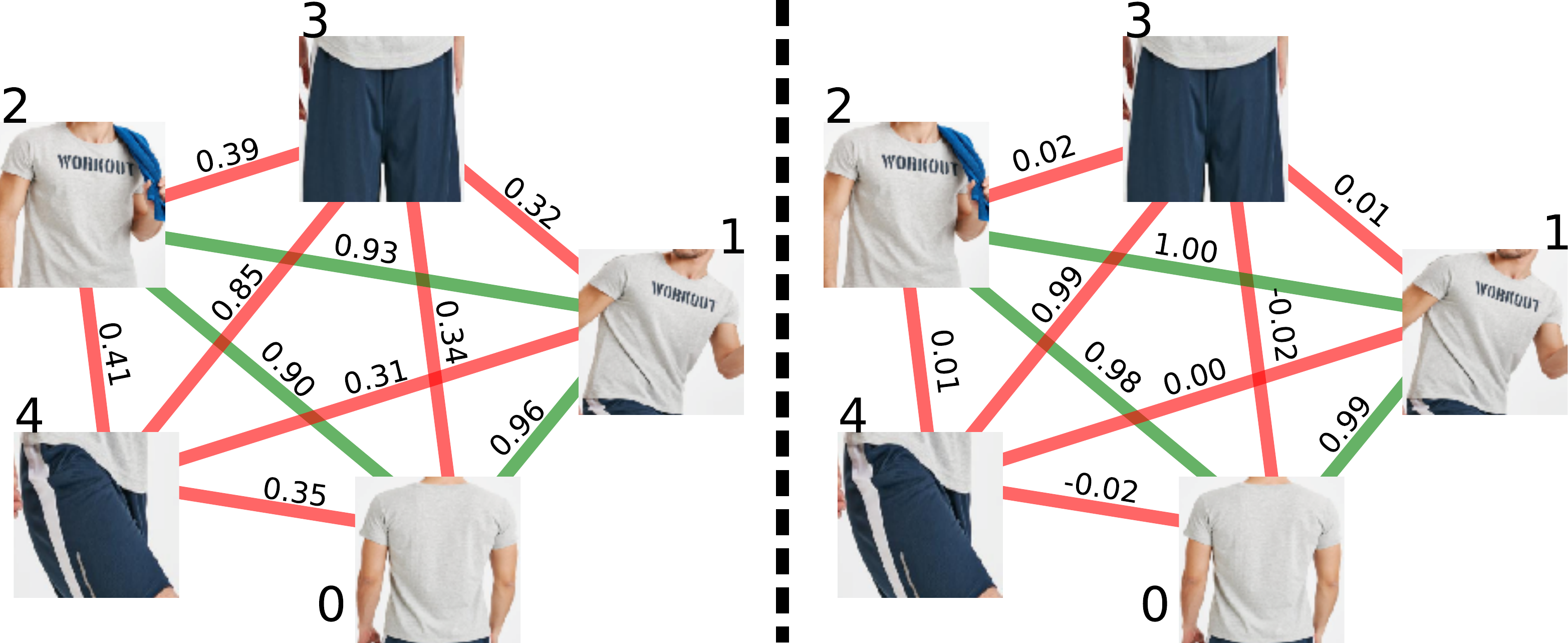}
  \caption{\small Cosine similarities between the features after the image model and the graph network respectively. After the node feature update, the similar items get closer to each other in the feature space, while the dissimilar items are pushed further away. The main bounding boxes are connected to each other with green edges. (best viewed in color)}
   \label{fig:last figure}
 \end{center}
\end{figure}

In Figure~\ref{fig:compare}, we display some qualitative results for the NG, PCFS and PDFS models. Moreover, in Figure~\ref{fig:last figure} it can be seen that after the node feature update the cosine similarities of node features are getting higher.

\begin{figure}[t]
  \centering
 \begin{subfigure}{.9\textwidth}
  \centering
  \includegraphics[width=.9\textwidth]{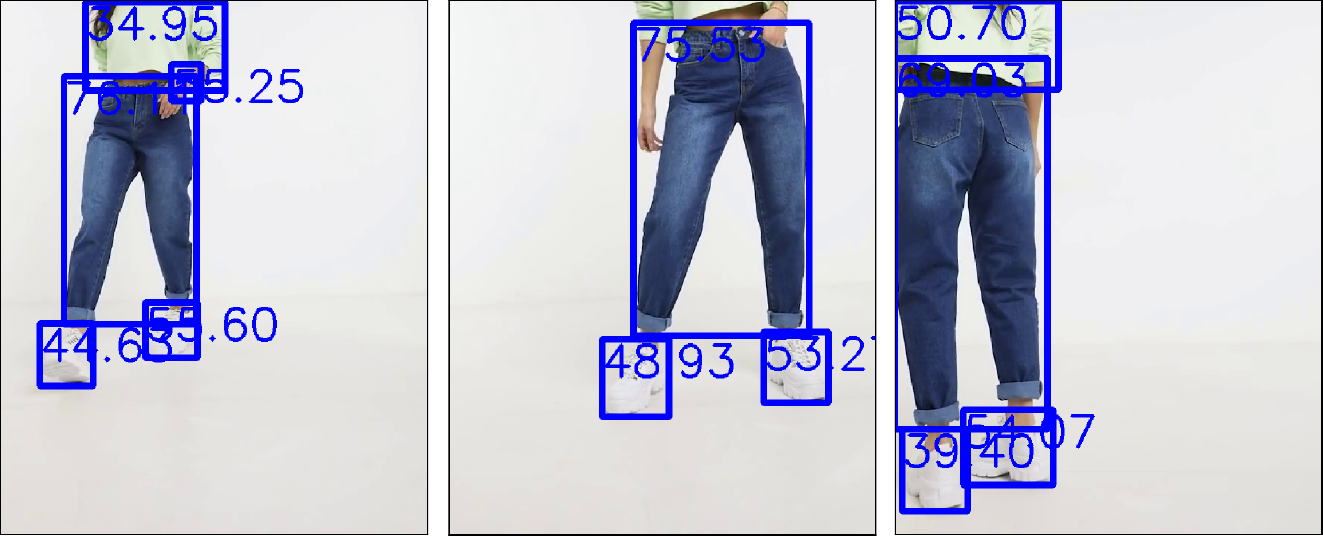}
  \caption{\textit{Mom jeans in blue}}
  \label{fig:video1}
\end{subfigure}%
\\
\begin{subfigure}{.9\textwidth}
  \centering
  \includegraphics[width=.9\textwidth]{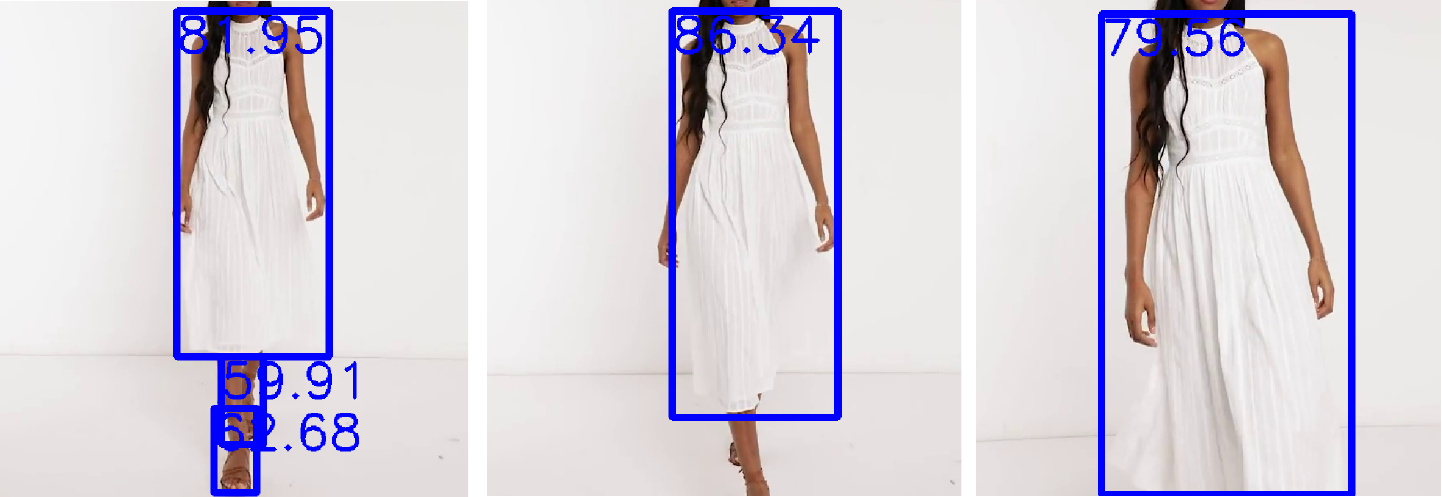}
  \caption{\textit{Dress with lace inserts in white}}
  \label{fig:video2}
\end{subfigure}%
\\
\begin{subfigure}{.9\textwidth}
  \centering
  \includegraphics[width=.9\textwidth]{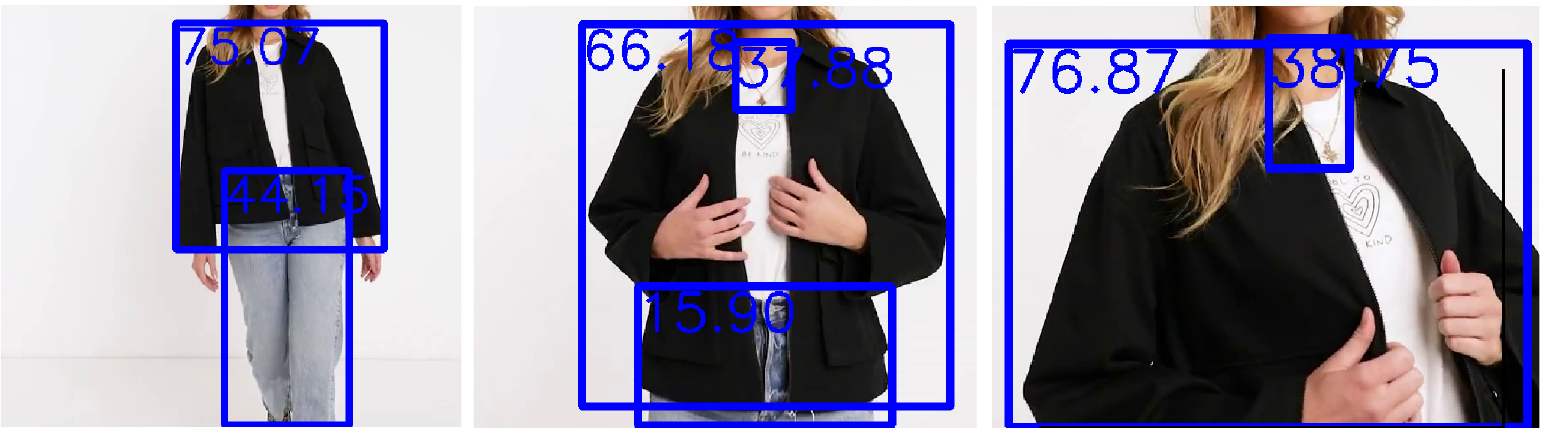}
  \caption{\textit{Shacket in black}}
  \label{fig:video3}
\end{subfigure}
\caption{ Evaluation of the main product detection on video frames. The product titles are written under the subfigure. In all frames, the bounding box that has the highest score is the main bounding box.}
\label{fig:video}
\end{figure}

\begin{table}[!tb]
\centering
\caption{Performance comparison of the baselines and graph-based approaches in the gallery-only setup, where no text input is provided}
\begin{tabular}{ccc|cccc}
Train               & Test               & Models      & P@1           & R@1           & mAP           & Prod. acc.    \\ \hline
\multirow{10}{*}{1} & \multirow{5}{*}{1} & Contrastive & 79.8          & 28.0          & 89.6          & 56.9          \\
                    &                    & NG          & 89.1          & 27.4          & 83.4          & 21.2          \\
                    &                    & ICFS        & \textbf{91.4} & \textbf{29.5} & 89.1          & 37.4          \\
                    &                    & PCFS        & 88.6          & 28.1          & \textbf{91.2} & 65.1          \\
                    &                    & PDFS        & 83.0          & 28.4          & 88.9          & \textbf{67.9} \\ \cline{2-7} 
                    & \multirow{5}{*}{2} & Contrastive & 81.6          & 39.3          & 91.6          & 65.1          \\
                    &                    & NG          & 80.5          & 38.3          & 84.2          & 37.8          \\
                    &                    & ICFS        & \textbf{88.4} & \textbf{41.4} & \textbf{90.4} & 51.2          \\
                    &                    & PCFS        & 81.3          & 39.0          & 89.3          & 67.6          \\
                    &                    & PDFS        & 81.8          & 39.2          & 89.6          & \textbf{71.3} \\ \hline
\multirow{10}{*}{2} & \multirow{5}{*}{1} & Contrastive & 54.0          & 19.5          & 74.2          & 20.7          \\
                    &                    & NG          & 84.1          & 26.1          & 78.5          & 21.8          \\
                    &                    & ICFS        & \textbf{87.8} & \textbf{27.2} & 79.5          & 24.8          \\
                    &                    & PCFS        & 77.4          & 24.5          & 73.4          & 21.3          \\
                    &                    & PDFS        & 80.1          & 25.0          & \textbf{80.6} & \textbf{30.4} \\ \cline{2-7} 
                    & \multirow{5}{*}{2} & Contrastive & 71.7          & 35.7          & 87.4          & 56.2          \\
                    &                    & NG          & 89.9          & 42.0          & 91.8          & 57.1          \\
                    &                    & ICFS        & 89.9          & 42.1          & 90.9          & 44.9          \\
                    &                    & PCFS        & 84.4          & 39.5          & 83.0          & 54.3          \\
                    &                    & PDFS        & \textbf{93.3} & \textbf{42.9} & \textbf{93.7} & \textbf{72.4}
\end{tabular}
\label{tab:results2}
\end{table}

\subsection{Gallery-only Main Product Detection}
In Table~\ref{tab:results2}, we evaluate the setup which we call gallery-only main product detection. In this setup, we take the best models from previous experiments and re-evaluate them while setting all input text embeddings to zero. This setup is an important indicator to evaluate models when they are deployed in the wild where product titles or descriptions will not always be available. It can be seen that the failure rate of the baseline approaches is much higher than the graph-based approaches. PCFS and PDFS models also yield better results than the ICSF model. This can be attributed to the fact that the graph-based models are able to enforce consistency between the bounding boxes thanks to the graph formulation, whereas the other methods show more dependency on the text, and fail in the case where no text input is provided. We attribute the relative high performance of the contrastive model to being biased to selecting the biggest bounding boxes as main products. The margin between the proposed and baselines approaches gets larger when the graph size increases, as can be seen in Figure~\ref{fig:curve2}.

Finally, as an additional illustration, in Figure~\ref{fig:video} we show that our method can be used to detect the main product in videos, by considering several frames from the video. The video frames are taken from products that are being sold in a website of a fashion retailer. In this website, along with the images, titles and descriptions of a product, a video of a model wearing the item is available to customers. We evaluated our PDFS model by randomly sampling 3 frames of a video. After running the fashion detector on these frames, the bounding boxes are input to the main product detection model along with the product title. In the figure, it can be seen that the main product detection model successfully assigns the highest scores to the main items compared to other items. This example shows that the proposed method here for product detection in gallery images can potentially also be used for detection of main products in fashion videos.

\subsection{Main Bounding Box Detection Dataset (MBBDD)}
We also train and evaluate the baselines and our models on MBBDD. We use the same models and hyperparameters that we used for the previous experiments. The results can be seen in Table~\ref{tab:results3}. Since the average number of positive bounding boxes per product is 1.02, the R@1 metric is much higher compared to results on our datasets. Again, especially in the product accuracy metric, graph based models achieve higher results than baselines. We do not display the results of the ICSF model since the products of the dataset are represented by single images.

\begin{table}[!tb]
\centering
\caption{Performance comparison of the baselines and graph-based approaches on MBBDD}
\begin{tabular}{c|cccc}
Models      & P@1           & R@1           & mAP           & Prod. acc.    \\ \hline
Contrastive & 95.5          & 94.5          & 97.7          & 88.4          \\
NG          & 96.3          & 95.3          & 98.1          & 91.1          \\
PCFS        & \textbf{96.5} & \textbf{95.5} & \textbf{98.2} & 94.8          \\
PDFS        & 96.4          & 95.4          & 98.1          & \textbf{94.8}
\end{tabular}
\label{tab:results3}
\end{table}

\section{Conclusions}
\label{sec:conclusions}
In this work, we propose a new approach for main product detection that incorporates a graph neural network to capture the relationships between all the detected products in a fashion product image gallery. We empirically demonstrate that the graph-based approaches surpass the baselines which do not take the context of product images into account with gains of 6-12 points. If we consider the more challenging  \emph{Gallery-only Main Product Detection} we show that using graphs can result in gains of up to 50 points when comparing to the same network without graphs. Moreover, with this work, we put a focus on the main product detection, a crucial but often overlooked task, that has received less attention from the research community due to its more application oriented structure.

\begin{acknowledgements}
This work was supported by the Spanish projects PID2019-104174GB-I00 and RTI2018-102285-A-I00, the Industrial Doctorate Grant 2016 DI 039 of the Ministry of Economy and Knowledge of the Generalitat de Catalunya, and its CERCA Program, and the Ramón y Cajal grant RYC2019-027020-I.
\end{acknowledgements}

\bibliographystyle{spmpsci}
\bibliography{longstrings,egbib}

\end{document}